# FedEPA: Enhancing Personalization and Modality Alignment in Multimodal Federated Learning

Yu Zhang, Qingfeng Du[✉], and Jiaqi Lv

School of Computer Science and Technology, Tongji University, Shanghai, China
`du_cloud@tongji.edu.cn`

**Abstract.** Federated Learning (FL) enables decentralized model training across multiple parties while preserving privacy. However, most FL systems assume clients hold only unimodal data, limiting their real-world applicability, as institutions often possess multimodal data. Moreover, the lack of labeled data further constrains the performance of most FL methods. In this work, we propose FedEPA, a novel FL framework for multimodal learning. FedEPA employs a personalized local model aggregation strategy that leverages labeled data on clients to learn personalized aggregation weights, thereby alleviating the impact of data heterogeneity. We also propose an unsupervised modality alignment strategy that works effectively with limited labeled data. Specifically, we decompose multimodal features into aligned features and context features. We then employ contrastive learning to align the aligned features across modalities, ensure the independence between aligned features and context features within each modality, and promote the diversity of context features. A multimodal feature fusion strategy is introduced to obtain a joint embedding. The experimental results show that FedEPA significantly outperforms existing FL methods in multimodal classification tasks under limited labeled data conditions.

## 1 Introduction

With the advancement of the information industry, organizations continuously collect and generate diverse data in their daily operations. Centralized machine learning systems, where data is aggregated on a single server to train generic models, face significant challenges such as data silos and privacy concerns. Federated Learning (FL) [1], a decentralized machine learning paradigm, enables multiple participants to collaboratively train models without sharing their raw data, offering a promising solution. However, existing FL methods typically assume clients hold only unimodal data, limiting their effectiveness in real-world scenarios where multimodal data is prevalent.

Recent studies on multimodal learning have demonstrated that leveraging complementary information across modalities significantly improves both performance and robustness compared to unimodal approaches, such as in weather forecasting [2], rumor detection [3], and medical diagnosis [4]. In the context of multimodal FL, integrating multimodal data presents unique opportunities to improve FL systems. Yet, notable challenges persist, including (1) personalizing models to address multimodal heterogeneity across clients, and (2) aligning multimodal features with limited labeled data.

To address the performance degradation caused by heterogeneous data distributions across clients, several studies have explored Personalized Federated Learning (PFL) approaches. For instance, FedPer [5] enhances model adaptability through personalized layers, and FedMBridge [6] employs a generative meta-learner to optimize client-specific parameters. However, these methods primarily focus on unimodal settings and lack effective mechanisms to handle heterogeneity in multimodal scenarios. In real-world FL systems, obtaining sufficient labeled data from clients is often challenging [7], further exacerbating the difficulty of aligning multimodal features, as different modalities may have varying annotation costs or availability. This limitation further complicates multimodal feature alignment. To mitigate this, SemiFL [8] enhances training by generating pseudo-labels, while $(FL)^2$ [9] employs adaptive thresholding to select unlabeled data for regularization, thereby improving learning performance. These approaches rely on selecting unlabeled data, which can easily introduce biases due to the unrepresentativeness of the data, such as favoring certain classes. Aligning multimodal features under limited supervision remains a significant challenge.

In this paper, we propose FedEPA, a novel multimodal FL framework that enhances model personalization to mitigate the impact of data heterogeneity across clients while improving multimodal feature alignment under limited labeled data. First, we introduce a personalized local aggregation strategy, which employs labeled client data to compute personalized aggregation weights when global model parameters are distributed from the server to initialize client models in each communication round. This approach enables clients to adaptively adjust the contribution of global parameters to their local models, enhancing model personalization. Second, we propose an unsupervised multi-modal feature alignment method based on feature decomposition. Specifically, we decompose the features of a sample into aligned features, which capture modality-shared semantics, and context features, which encapsulate modality-specific information. We enhance multimodal alignment by (1) improving the consistency of aligned features across modalities of the same data to extract shared information, (2) separating aligned features and context features within a modality to ensure their semantic independence, and (3) maintaining diversity among context features across different samples and modalities to prevent convergence of task-irrelevant information. Additionally, we design a multimodal fusion strategy to obtain a joint embedding representation, improving performance in multimodal classification tasks. Our contributions are as follows:

- We propose a personalized weighted local aggregation strategy, adjusting the global model to match each client's data distribution, effectively handling heterogeneity.
- We design an unsupervised feature alignment method that decomposes multimodal features and optimizes their representation using unlabeled data, paired with a fusion strategy to enhance multimodal classification performance.
- Experiments on three public datasets demonstrate the efficiency of FedEPA in multimodal classification scenarios with limited label data.

## 2    Related Work

**Multimodal Federated Learning (MMFL)**: MMFL enables collaborative training using multimodal data distributed across clients. MM-FedAvg [10] employs an autoencoder to align multimodal data for improved feature fusion, FedCMR [11] learns shared latent subspaces for cross-modal retrieval, and FedMEKT [12] uses knowledge distillation with proxy datasets to transfer multimodal embeddings. However, they struggle with heterogeneous data distribution across clients. FedEPA introduces personalized aggregation weights to adapt client models to their local multimodal distributions.

**Multimodal Contrastive Learning**: Contrastive learning has shown promise in multimodal feature alignment. CLIP [13] aligns vision and text using natural language, CMC [14] extracts shared features by maximizing mutual information, and FOCAL [15] decomposes data into orthogonal spaces. The lack of personalization limits their applicability in MMFL. FedEPA combines contrastive learning with personalized aggregation to improve multimodal alignment under heterogeneity and label scarcity.

## 3      Preliminaries

In MMFL framework proposed in our work, the client set is denoted as $\mathcal{N} = \{1, 2, \ldots N\}$, where each client holds private multimodal data. Under the coordination of a central server $G$, a shared global model and $N$ client local models are trained to perform multimodal classification tasks.

For a client $i \in \mathcal{N}$, its local dataset $D_i$ contains labeled data $D_i^l$ and unlabeled data $D_i^u$. Each client data comprises $M$ modalities, denoted as $\mathcal{M} = \{1, 2, \ldots, M\}$. A sample $x_{ij}$ in $D_i$, encompasses multiple modalities, expressed as $x_{ij} = \{x_{ij}^1, x_{ij}^2, \ldots, x_{ij}^M\}$, with labeled data associated with a label $y_{ij}$. Both the server and clients maintain a model $F$, which consists of a set of modality-specific encoders $\{\mathcal{E}_m\}_{m \in \mathcal{M}}$ and a classifier $\mathcal{C}$. The federated learning process comprises $T$ rounds of communication, where each client performs $R$ local iterations with a batch size of $\mathcal{B}$.

## 4      Methodology

In this section, we introduce our proposed FedEPA framework. As illustrated in Fig. 1, FedEPA comprises three main components: (a) Personalized Weighted Local Aggregation, (b) Unsupervised Modality Alignment, and (c) Multimodal Feature Fusion. We will introduce them in Sections 4.1, 4.2, and 4.3, respectively.

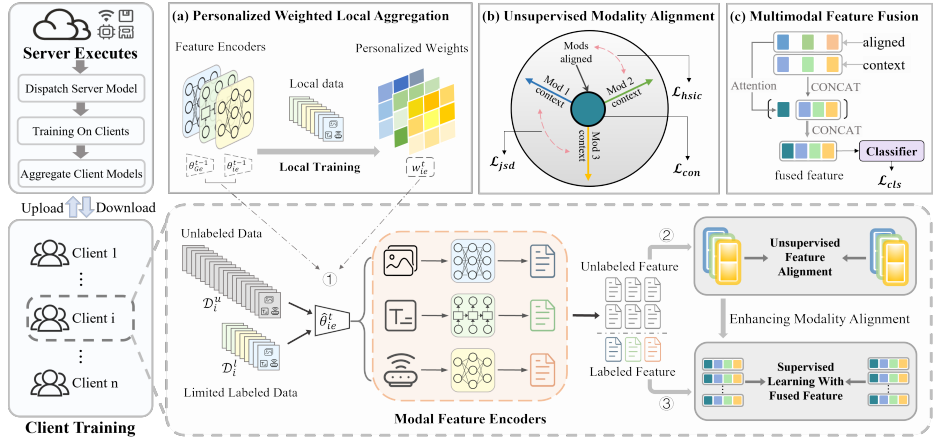

**Fig. 1.** The illustration of the FedEPA framework with three modalities of data as an example on each client. ① update client model with personalized weights; ② unsupervised multimodal feature alignment training; ③ supervised classification based on multimodal feature fusion.

### 4.1    Personalized Weighted Local Aggregation

In this section, we present a personalization aggregation strategy for clients. In traditional FL frameworks, at each server iteration $t$, the parameters $\theta_G^{t-1}$ of the global model $F_G$, aggregated from the parameters of participating clients in the preceding round, are distributed to each client. $\theta_G^{t-1}$ serves as the initial parameters for $\hat{\theta}_i^t$ of the model $F_i$ on client $i$ in the current round, as formalized by the expression:

$$\hat{\theta}_i^t \leftarrow \theta_G^{t-1} \tag{1}$$

However, due to the heterogeneous data distributions across clients, directly initializing client models with the global model by Equation (1) may perform poorly on certain local datasets, further degrading the overall model's effectiveness after aggregation.

To address this problem, we propose a personalized weight-based local aggregation strategy optimized for multimodal encoders. We represent the global model parameters as $\theta_G^{t-1} = [\theta_{Ge}^{t-1}, \theta_{Gc}^{t-1}]$, which denote the parameters of the multimodal feature encoders and the classifier, respectively. Similarly, the local model parameters of client $i$ are denoted as $\theta_i^{t-1} = [\theta_{ie}^{t-1}, \theta_{ic}^{t-1}]$.

Inspired by [16, 17], we introduce a learnable personalized weight $w_i^t$, optimized using labeled local data to adaptively control the contribution of the global model to the local model update. Since the classifier parameters $\theta_{Gc}^{t-1}$ undergo aggregation across multiple clients and tend to be more robust, we freeze them during personalized local training. Thus, the weight vector $w_i^t = [w_{ie}^t, w_{ic}^t]$ can be simplified as $[w_{ie}^t, \mathbf{1}_c]$, where $\mathbf{1}_c$ is an all-ones matrix matching the shape of $\theta_{Gc}^{t-1}$. The personalized local aggregation process can be formulated as follows:

$$\hat{\theta}_i^t \leftarrow [\theta_{ie}^{t-1} + (\theta_{Ge}^{t-1} - \theta_{ie}^{t-1}) \odot w_{ie}^t, \theta_{Gc}^{t-1}] \tag{2}$$

where $\odot$ denotes element-wise multiplication. Initially, $w_{ie}^{t-1}$ is an all-ones matrix, meaning that, without personalized aggregation, Equation (2) reduces to Equation (1).

We then optimize $w_{ie}^t$ using the labeled local dataset $D_i^l$ through supervised learning to minimize the following loss function:

$$\mathcal{L}(\theta_e^p) = \frac{1}{|D_i^l|} \sum_{j=1}^{|D_i^l|} l\left(F_i(\theta_e^p, x_{ij}), y_{ij}\right) \tag{3}$$

where $l$ represents the cross-entropy loss function, and $\theta_e^p$ denotes the temporary local model parameters on client $i$. Previous studies [18, 19] suggest that during backpropagation, neurons with larger gradients play a crucial role in model learning. Based on this observation, we define the update rule for $w_{ie}^t$ as:

$$w_{ie}^t \leftarrow clamp\left(w_{ie}^t - \nabla_{\theta_{ie}^p}(\theta_{Ge}^{t-1} - \theta_{ie}^p)\right) \tag{4}$$

where $\nabla_{\theta_{ie}^p}$ represents the gradient of the loss function with respect to $\theta_e^p$ during personalized training, and $clamp(\cdot)$ ensures that the weights remain within the interval $[0,1]$ (i.e., between 0 and 1). Through iterative updates, we obtain the personalized local model $\hat{\theta}_i^t$.

## 4.2 Unsupervised Modality Alignment

In this section, we present an unsupervised alignment strategy for multimodal data, designed to capture shared information across modalities while reducing the impact of noise. For a sample $j$ from client $i$ in modality $m$, we map its input data $x_{ij}^m$ to a feature representation $z_{ij}^m = \mathcal{E}_{im}(x_{ij}^m)$ via an encoder. Assuming the feature dimension is $2d$, we decompose $z_{ij}^m$ as follows:

$$z_{ij}^{ma} = z_{ij}^m[0:d], \; z_{ij}^{mc} = z_{ij}^m[d:2d] \tag{5}$$

where $z_{ij}^{ma}$ represents the aligned features, which capture the core semantic information of modality $m$, while $z_{ij}^{mc}$ represents the context features of modality $m$, which encode auxiliary or noisy information.

**Consistency Constraint for Aligned Features.** To encourage alignment across different modalities, we apply an unsupervised contrastive learning objective using unlabeled samples to enhance the consistency of aligned features $z_{ij}^{ma}$, $z_{ij}^{na}$ between modalities $m$ and $n$ for the same sample $j$. Given a batch $\mathcal{B}$ from $D_i^u$, the learning objective is defined as:

$$\mathcal{L}_{con} = -\sum_{m,n \in \mathcal{M}, m \neq n} log\left(\frac{exp(sim(z_{ij}^{ma}, z_{ij}^{na})/\tau)}{\sum_{j,k \in B, j \neq k} exp(sim(z_{ij}^{ma}, z_{ik}^{na})/\tau)}\right) \tag{6}$$

where $sim(\cdot, \cdot)$ denotes cosine similarity, $exp(\cdot)$ is the exponential function, and $\tau$ is a temperature parameter. By minimizing $\mathcal{L}_{con}$, we encourage the aligned features from different modalities of the same sample to be close in the feature space, thereby enhancing multimodal feature alignment.

**Independence Constraint Between Aligned and Context Features.** To ensure the aligned features $z_{ij}^{ma}$ and the context features $z_{ij}^{mc}$ of modality $m$ for sample $j$ capture relatively independent semantic information, we employ the Hilbert-Schmidt Independence Criterion (HSIC) [20] to constrain their statistical dependence. Given a batch $\mathcal{B}$ from $D_i^u$, the independence loss is defined as:

$$\mathcal{L}_{hsic} = \sum_{j \in \mathcal{B}} \sum_{m \in \mathcal{M}} HSIC\left(z_{ij}^{ma}, z_{ij}^{mc}\right) \tag{7}$$

where $HSIC(P, Q) = \frac{1}{(d-1)^2} Tr(K_P H K_Q H)$, $Tr$ denotes the trace of matrix, $K_P$ and $K_Q$ are kernel matrices (Gaussian kernels in our work) for feature matrices $P$ and $Q$, and $H = I - \frac{1}{d}$.

By minimizing $\mathcal{L}_{hsic}$, we reduce the dependency between $z_{ij}^{ma}$ and $z_{ij}^{mc}$, ensuring that aligned features focus on core information while context features retain independent environmental and noise-related information of the modality.

**Diversity Constraint for Context Features.** To ensure diversity among the context features $z_{ij}^{mc}$ and $z_{ik}^{nc}$ across different samples and modalities, preventing convergence toward task-irrelevant information, we introduce the Jensen-Shannon Divergence (JSD) as a constraint. Given a batch $\mathcal{B}$ from $D_i^u$, the diversity loss is defined as:

$$\mathcal{L}_{jsd} = \sum_{j,k \in \mathcal{B}} \sum_{m,n \in M, m \neq n} JSD\left(z_{ij}^{mc}, z_{ik}^{nc}\right) \tag{8}$$

where $JSD(P, Q) = \frac{1}{2}(KL(P||\frac{P+Q}{2}) + KL(\frac{P+Q}{2}||Q))$, with $P$ and $Q$ representing the distribution estimates of $z_{ij}^{mc}$ and $z_{ik}^{nc}$, respectively, and $KL(\cdot)$ denoting the Kullback-Leibler divergence, which measures the asymmetric difference between two distributions. By minimizing $\mathcal{L}_{jsd}$, we ensure that context features remain diverse across different modalities and samples, preserving the richness of multimodal data.

The total training objective for unsupervised contrastive feature alignment is $\mathcal{L}_{align}$, where $\lambda_1$ and $\lambda_2$ are hyperparameters balancing the loss terms:

$$\mathcal{L}_{align} = \mathcal{L}_{con} + \lambda_1 \mathcal{L}_{hsic} + \lambda_2 \mathcal{L}_{jsd} \tag{9}$$

### 4.3 Multimodal Feature Fusion

The alignment strategy introduced in Section 4.2 helps extract complementary information across modalities while reducing noise interference. Building upon this strategy, we design a multimodal feature fusion strategy to integrate information dynamically. Specifically, we employ a self-attention mechanism [21] to capture cross-modal correlations and evaluate the relative importance across different modalities in the fusion process. For the aligned feature $z_{ij}^{ma}$ of each modality $m$, we compute their Query, Key, and Value matrices as follows:

$$q_{ij}^m = W_q z_{ij}^{ma}, k_{ij}^m = W_k z_{ij}^{ma}, v_{ij}^m = W_v z_{ij}^{ma} \tag{10}$$

where $W_q$, $W_k$, and $W_v$ are learnable transformation matrices. The attention weight matrix between modality $m$ and $n$ is given by:

$$a_{ij}^{mn} = softmax\left(\frac{q_{ij}^m (k_{ij}^n)^T}{\sqrt{d}}\right) \tag{11}$$

For each sample $j$ on client $i$, given its multimodal features $\{z_{ij}^m, m \in \mathcal{M}\}$, the fusion strategy is defined as:

$$f_{ij} = CONCAT\left(\sum_{m,n \in M} a_{ij}^{mn} v_{ij}^n, \{z_{ij}^{mc}, m \in \mathcal{M}\}\right) \tag{12}$$

where $CONCAT(\cdot)$ concatenates the core information with the contextual features of each modality. We retain $z_{ij}^{mc}$ to fully leverage the diversity of multimodal data, thereby enhancing the robustness and generalization ability of the classifier.

For the multimodal classification task in this study, the fused features are fed into the classifier $\mathcal{C}$, and the model is trained using a supervised cross-entropy loss:

$$\mathcal{L}_{cls} = \frac{1}{|D_i^l|} \sum_{j=1}^{|D_i^l|} l(\mathcal{C}(f_{ij}), y_{ij}) \tag{13}$$

The overall pseudocode for the FedEPA process is presented in Appendix C.

## 5    Experiments

In this section, we begin by introducing the datasets and experimental settings used in our study. Then, we compare FedEPA with other state-of-the-art methods on three real-world datasets, including a t-SNE visualization for interpretability. We also demonstrate the effectiveness of the multimodal feature fusion strategy. Additionally, we evaluate the robustness of FedEPA under different label proportions and further analyze its stability in non-IID environments.

### 5.1    Datasets

We conduct experiments on three multimodal classification datasets: MGCD [22], Weibo [3], and UTD-MHAD [23].

**MGCD** contains 8,000 multimodal cloud samples, including cloud images and meteorological data, and is categorized into seven types: cumulus, altocumulus, cirrus, clearsky, stratocumulus, cumulonimbus, and mixed.

**Weibo** is collected from the official Weibo rumor debunking system, containing 9,528 posts related to rumor detection that include both images and text.

**UTD-MHAD** consists of 27 different actions performed by 8 subjects, with each action repeated 4 times. After removing 3 corrupted sequences, 861 sequences remain. We use depth images, skeleton joints, and inertial sensor data for action recognition.

### 5.2    Experimental Settings

**Compared Methods**. To evaluate the performance of FedEPA, we compare it against several state-of-the-art methods, including the classic FL algorithms FedAvg [0] and FedProx [24], as well as PFL methods such as FedPer [5], FedRep [25], FedALA [17], FedDBE [26], FedAS [27], and MMFL methods such as FedCMR [11] and FDARN [28]. For a more detailed description, please refer to Appendix A.

**Metrics**. We evaluate the performance of FedEPA in multimodal classification using three metrics: Overall Accuracy (OA), calculated as the ratio of correctly classified samples to total samples; Balanced Accuracy (BA), obtained as the average recall across all classes; and F1-Score (F1), the harmonic mean of precision and recall. The calculation details for these metrics are provided in Appendix B.

**Implementation Details.** The implementation of FedEPA is based on PyTorch, and all experiments are conducted on an NVIDIA GeForce RTX 4090 24GB GPU. All experiments utilize a training configuration with a learning rate of 0.0005 and a batch size of 32. Following prior work [24, 26], we partition the client data into a 4:1 ratio for training and test sets. To simulate a realistic federated learning scenario with limited labeled data, we designate 20% of the training set as labeled data. As we focus on multimodal federated learning, the design of encoders is kept lightweight. For multimodal data processing, image data is encoded by a two-layer CNN, sequential data with temporal characteristics is processed by an LSTM for feature extraction, and tabular data is encoded by an MLP. Finally, a fully connected layer serves as the classifier.

## 5.3    Main Results

In general, the following observations presented in Table 1 demonstrate that our proposed FedEPA method achieves the best performance across all three datasets.

**Table 1.** Performance (%) comparison on MGCD, Weibo, and UTD-MHAD. The best result is shown in bold; the second best is marked with underline; w/o means without for ablation study.

| Method | Dataset | MGCD | | | Weibo | | | UTD-MHAD | | |
|---|---|---|---|---|---|---|---|---|---|---|
| | Metric | OA | BA | F1 | OA | BA | F1 | OA | BA | F1 |
| FedAvg | | 65.91 | 64.37 | 66.00 | 76.78 | 75.68 | 76.04 | 40.51 | 41.86 | 37.66 |
| FedProx | | 68.32 | 66.01 | 67.33 | 77.25 | 78.16 | 77.81 | 42.44 | 44.03 | 38.69 |
| FedPer | | 59.07 | 55.43 | 56.95 | 78.14 | 76.51 | 77.32 | 27.65 | 29.26 | 23.07 |
| FedRep | | 51.33 | 46.42 | 44.54 | 75.96 | 74.77 | 75.54 | 53.41 | 54.34 | 58.97 |
| FedALA | | 66.46 | 63.66 | 64.12 | 77.12 | 78.58 | 77.06 | 63.45 | 62.11 | 62.64 |
| FedDBE | | 69.30 | 67.13 | 68.31 | 79.41 | 77.16 | 78.75 | 38.26 | 40.71 | 34.24 |
| FedAS | | 66.99 | 63.31 | 63.09 | 77.64 | 75.63 | 77.09 | 36.01 | 37.47 | 34.35 |
| FedCMR | | 67.31 | 65.34 | 67.14 | 80.00 | 78.96 | 79.86 | 45.80 | 47.83 | 41.24 |
| FDARN | | 57.34 | 54.58 | 56.33 | 75.94 | 74.42 | 75.72 | 48.30 | 42.70 | 45.39 |
| Ablation | FedEPA-w/o PA | 78.24 | 77.04 | 71.04 | 75.53 | 73.96 | 75.41 | 62.30 | 63.26 | 59.88 |
| | FedEPA-w/o UA | <u>87.52</u> | <u>86.90</u> | <u>87.44</u> | <u>81.76</u> | <u>80.50</u> | <u>81.56</u> | <u>76.21</u> | <u>77.89</u> | <u>76.21</u> |
| Ours | FedEPA | **91.20** | **90.78** | **91.24** | **83.03** | **82.35** | **82.45** | **83.92** | **84.76** | **83.66** |

The experimental results highlight the superiority of FedEPA over other compared methods. On the MGCD and UTD-MHAD datasets, FedEPA achieves performance gains exceeding 30%. Especially on the UTD-MHAD dataset, its small dataset size and numerous categories exacerbate the difficulty of modality alignment with limited labeled data and hinder many methods from effectively utilizing multimodal information. For example, FedPer and FedDBE perform well on the Weibo and MGCD datasets but experience a significant drop in performance on the UTD-MHAD dataset, suggesting they fail to effectively address modality alignment issues. In contrast, FedEPA employs an unsupervised modality alignment strategy (UA) to improve feature representation by leveraging unlabeled data, enhancing its robustness and effectiveness. Moreover, some PFL methods (e.g., FedPer and FedRep) perform even worse than FedAvg on the MGCD dataset, primarily due to their inadequate optimization for heterogeneity in multimodal scenarios. In comparison, FedEPA effectively overcomes this challenge with its personalized weighted local aggregation strategy (PA), leading to comprehensive performance improvements and highlighting its superiority in tackling heterogeneity. Furthermore, we conducted two ablation studies to validate the effectiveness of PA and UA. The results show that eliminating UA reduces performance across datasets, emphasizing its critical role in improving feature representation in multimodal, label-scarce contexts. Similarly, excluding PA significantly impacts the results, highlighting its importance in managing client data heterogeneity in multimodal federated learning.

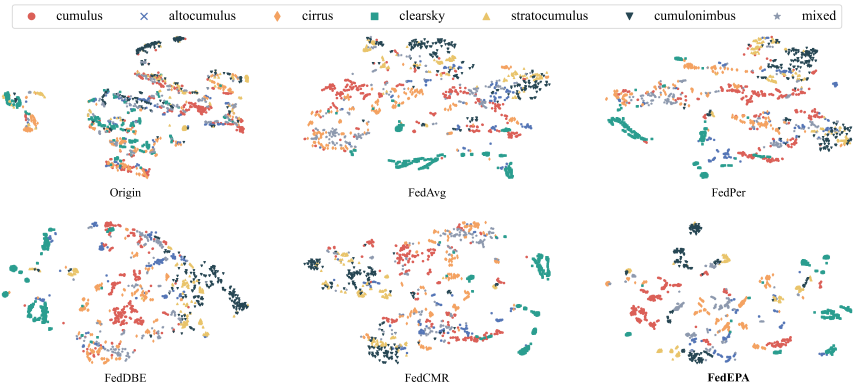

**Fig. 2.** The t-SNE visualization of selected embeddings.

We then analyze the feature representation ability of different methods by applying t-SNE to visualize embeddings on the MGCD dataset before classification. The visualization results in Fig. 2 reveal that raw features exhibit blurry class boundaries, with samples heavily mixed together, making class separation difficult. Additionally, FedPer performs worse than FedAvg in some classes (e.g., cumulonimbus and stratocumulus), increasing overlap and indicating that traditional PFL methods may struggle with multimodal feature representation. Conversely, FedEPA achieves a more apparent class separation, with samples forming compact, distinct clusters in t-SNE visualization. This demonstrates the effectiveness of its personalized learning strategy and unsupervised alignment mechanism in enhancing feature quality and model performance.

### 5.4 Multimodal Fusion Strategy Analysis

We evaluate the effectiveness of our proposed multimodal feature fusion strategy across three datasets and compare it against competitors that employ fusion techniques, such as Concat Fusion (directly concatenates the features of each modality into a single vector without decomposition), Aligned-Only Fusion (fuses only the aligned portions of each modality's features using attention mechanisms), and Add Fusion (directly sums the features of each modality element-wise without decomposition).

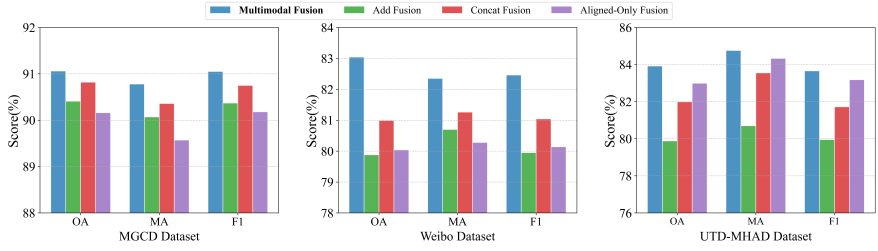

**Fig. 3.** Performance comparison of FedEPA under different fusion strategies.

As shown in Fig. 3, FedEPA with our fusion strategy consistently outperforms all competitors across all datasets, achieving performance improvements of up to 2%, which demonstrates the effectiveness of feature alignment and multimodal fusion strategy. Meanwhile, Aligned-Only Fusion performs even worse than Concat Fusion and Add Fusion on the Weibo and MGCD datasets, indicating that context features still contribute to classification tasks. Our fusion strategy effectively extracts and integrates this information, enhancing the model's robustness and overall performance.

## 5.5    Effect of Label Ratio

To evaluate the effectiveness of the unsupervised feature alignment module under varying label proportions, we conducted experiments with label ratios in the training set increasing from 10% to 100% in 10% increments on selected datasets.

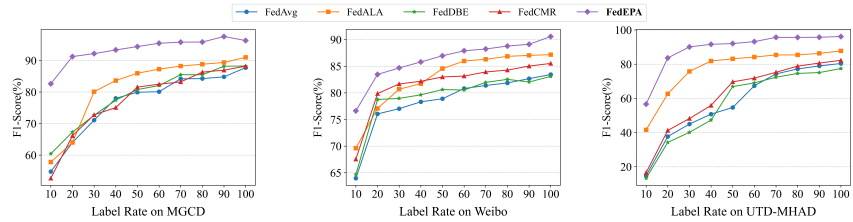

**Fig. 4.** Results of experiments with different ratios of labeled training data.

As shown in Fig. 4, the experimental results indicate that FedEPA consistently outperforms the four representative methods across different label proportions. Its robustness is particularly evident with 10% labeled training data, where it achieves performance gains of over 20% on the MGCD dataset, 15% on the UTD-MHAD dataset, and 10% on the Weibo dataset compared to other methods. This highlights the advantage of the unsupervised alignment strategy in handling label scarcity and leveraging unlabeled samples to enhance model performance. Notably, even with 100% labeled data, when unsupervised alignment is inactive, FedEPA still surpasses other methods, highlighting the benefits of the personalized aggregation strategy. An interesting observation on the MGCD dataset is that the performance of FedEPA with 100% labeled data is slightly lower than with 90%, suggesting that unsupervised alignment using partially labeled data can, as observed on the MGCD dataset, optimize feature representations better than fully supervised training. This further validates the potential of our strategy.

### 5.6    Non-IID Experiment

To evaluate the performance of FedEPA in non-IID (non-independent and identically distributed) scenarios, we conducted an experiment on the MGCD dataset, using the Dirichlet distribution to create heterogeneous data distributions across clients. By adjusting the Dirichlet distribution parameter $\beta$, we controlled the degree of data heterogeneity, where a smaller $\beta$ yields highly non-IID data among clients, while a larger $\beta$ approaches IID conditions. In this experiment, we set $\beta \in \{0.1, 0.5, 1.0, 10.0\}$ to cover varying heterogeneity levels from strong non-IID to near-IID scenarios.

**Table 2.** Overall accuracy (%) on the MGCD dataset under different non-IID settings.

| $\beta$ | FedAvg | FedProx | FedPer | FedALA | FedCMR | FedDBE | FedEPA |
|---------|--------|---------|--------|--------|--------|--------|--------|
| 0.1 | 54.43 | 60.56 | 46.02 | 30.65 | 52.93 | 56.74 | 89.13 |
| 0.5 | 56.94 | 62.12 | 44.96 | 37.99 | 55.87 | 58.53 | 89.37 |
| 1.0 | 60.50 | 66.24 | 53.26 | 52.59 | 64.70 | 66.03 | 90.64 |
| 10.0 | 63.82 | 67.49 | 58.63 | 64.85 | 65.37 | 68.93 | 91.05 |

As shown in Table 2, FedEPA consistently outperforms all compared methods under varying $\beta$ settings. Even when $\beta = 0.1$, FedEPA maintains stable performance with a decline of less than 2% compared to $\beta = 10.0$. This demonstrates the effectiveness of FedEPA in non-IID scenarios and highlights its robustness and broad applicability.

## 6      Conclusion

In this paper, we present FedEPA, a novel multimodal federated learning framework designed to address data heterogeneity and enhance multimodal alignment under limited labeled data. FedEPA improves model adaptability to heterogeneous data by optimizing client model aggregation using personalized weights. It also incorporates an unsupervised feature decomposition and alignment strategy to enhance multimodal feature alignment across modalities, as well as a multimodal feature fusion strategy to enhance robustness. Experiments show that FedEPA outperforms existing methods on three multimodal datasets, particularly in label-scarce scenarios.

**Acknowledgments.** This study was funded by the National Key Research and Development Program of China (No. 2022ZD0120300).

**Disclosure of Interests.** The authors have no competing interests to declare that are relevant to the content of this article.

## Appendix

### A.   Detailed Compared Methods

FedAvg [1]: Averages local model parameters from all clients to update the global model, ideal for distributed training.

FedProx [24]: Extends FedAvg with a regularization term to limit local-global model divergence, improving stability in heterogeneous settings.

FedPer [5]: Splits models into shared and personalized layers, aggregating only shared layers globally to handle task heterogeneity.

FedRep [25]: Balances global consistency and local adaptability by optimizing shared features and local classifiers alternately.

FedALA [17]: Adaptively aggregates the global and local models using an Adaptive Local Aggregation (ALA) module to initialize client models toward local objectives.

FedDBE [26]: Separates local-specific biases using Personalized Representation Bias Memory (PRBM) and uses Mean Regularization to ensure consistent, unbiased global representations.

FedCMR [11]: Extracts modality-invariant and specific features with encoders, ensuring orthogonality and optimizing global/private classifiers.

FedAS [27]: Addresses asynchronous update issues by localizing parameters and uses server-side aggregation to prioritize key client contributions.

FDARN [28]: Employs cross-modal retrieval to map global features to a local subspace, minimizing information loss in multimodal tasks.

## B.   Detailed Metrics

**Overall Accuracy (OA)**: OA measures the proportion of correct predictions across all classes, calculated as:

$$OA = \frac{TP+TN}{TP+TN+FP+FN} \tag{14}$$

where $TP$ denotes the number of true positives, $TN$ the number of true negatives, $FP$ the number of false positives, and $FN$ the number of false negatives.

**Balanced Accuracy (BA)**: BA offers a robust measure for multiclass tasks by averaging recall across $C$ classes, effectively capturing performance across diverse categories. It is calculated as:

$$BA = \frac{1}{C}\sum_{i=1}^{C}\frac{TP_i}{TP_i+FN_i} \tag{15}$$

**F1-Score (F1)**: F1 harmonizes precision and recall into a single metric, given by:

$$F1 = 2 \cdot \frac{Precision \cdot Recall}{Precision + Recall} \tag{16}$$

where $Precision = \frac{TP}{TP+FP}$ and $Recall = \frac{TP}{TP+FN}$. F1 balances precision and recall, providing a more holistic evaluation of model performance.

## C.  Pseudocode of FedEPA

---

**Algorithm 1: FedEPA**

---

**Input:** The client datasets $\{D_1, D_2, \ldots, D_N\}$, communication rounds $T$, epochs $R$
**Output:** The server model parameters $\theta_G^T$, clients model parameters $\{\theta_1^T, \theta_2^T, \ldots, \theta_N^T\}$

1:  **ServerExecutes:**
2:   Initialize $\theta_G^0$ randomly
3:   **for** t = 1 to T **do**
4:    │  Dispatch $\theta_G^{t-1}$ to all clients in $\mathcal{N}$          *// Dispatch server model*
5:    │  **for** each client $i \in \mathcal{N}$ in parallel **do**
6:    │  │  $\theta_i^t \leftarrow$ **ClientLocalTrainig**$(i, \theta_G^{t-1})$   *// Training on clients*
7:    │  **end for**
8:    │  $\theta_G^t \leftarrow \frac{1}{N}\sum_i^N \theta_i^t$             *// Aggregate client models*
9:   **end for**
10:  **return** $\theta_G^T$

11:  **ClientLocalTraining** $(\theta_G^{t-1})$**:**
12:   $\hat{\theta}_i^t \leftarrow \boldsymbol{PersonalAggregation}(\theta_G^{t-1}, \theta_i^{t-1}, D_i^l)$
13:   Initialize $\theta_i^t$ with $\hat{\theta}_i^t$
14:   **for** iteration $r = 1$ to $R$ **do**
15:    │  **for** batch data in $D_i^u$ **do**
16:    │  │  Update $\theta_{ie}^t$ according to **Equation (9)**
17:    │  **end for**
18:   **end for**
19:   **for** iteration $r = 1$ to $R$ **do**
20:    │  **for** batch data in $D_i^l$ **do**
21:    │  │  Get fused feature according to **Equation (11)**
22:    │  │  Compute $L_{cls}$ according to **Equation (12)**
23:    │  │  Update $\theta_i^t$ by minimizing $L_{cls}$
24:    │  **end for**
25:   **end for**
26:  **return** $\theta_i^t$

27:  **PersonalAggregation** $(\theta_i^{t-1}, \theta_G^{t-1}, D_i^l)$**:**
28:   Initialize $w_{ie}^t$ as an all-ones matrix with the same shape as $\theta_i^{t-1}$
29:   Initialize $\theta_i^t$ with $\theta_G^{t-1}$, initialize $\theta_i^p$ with $\theta_i^{t-1}$, and freeze $\theta_{ic}^p$
30:   **for** each $(x_{ij}, y_{ij})$ in $D_i^l$ **do**
31:    │  $\mathcal{L}(\theta_e^p) = \frac{1}{|D_i^l|}\sum_{j=1}^{|D_i^l|} l\left(F_i(\theta_e^p, x_{ij}), y_{ij}\right)$
32:    │  $w_{ie}^t = clamp(w_{ie}^t - \nabla_{\theta_{ie}^p}(\theta_{Ge}^{t-1} - \theta_e^p))$
33:   **end for**
34:   $\hat{\theta}_i^t \leftarrow [\theta_{Ge}^{t-1} + (\theta_{Ge}^{t-1} - \theta_{ie}^{t-1}) \odot w_{ie}^t, \theta_{Gc}^{t-1}]$
35:  **return** $\hat{\theta}_i^t$